\documentclass{clv2}
\usepackage{times}
\usepackage{url}
\usepackage{latexsym}
\usepackage{graphicx}
\usepackage{algorithm}
\usepackage[noend]{algpseudocode}
\usepackage{multirow}

\usepackage{amsmath}
\usepackage{amssymb}
\usepackage{bm}
\usepackage{balance}

\runningtitle{Context-Dependent Translation Selection Using Convolutional Neural Network}
\runningauthor{Zhaopeng Tu}

\begin{document}
\title{Context-Dependent Translation Selection Using Convolutional Neural Network}
\author{Zhaopeng Tu\thanks{Units 525-530, Core Building 2, 
Hong Kong Science Park, Shatin, Hong Kong. Email: tuzhaopeng@gmail.com}}
\affil{Huawei Technologies Noah's Ark Lab, Hong Kong}

\author{Baotian Hu}
\affil{Harbin Institute of Technology, Shenzhen Graduate School}

\author{Zhengdong Lu}
\affil{Huawei Technologies Noah's Ark Lab, Hong Kong}

\author{Hang Li}
\affil{Huawei Technologies Noah's Ark Lab, Hong Kong}

%\sndaff ~~~~~~~~~ Baotian Hu\fstaff ~~~~~~~~~ Zhengdong Lu\sndaff ~~~~~~~~~ Hang Li\sndaff
%\\
%\begin{tabular}{cc}
%{\sndaff Noah’s Ark Lab}  &  {\fstaff Department of Computer Science \& Technology}   \\
%{Huawei Technologies Co. Ltd.}                   &       {Harbin Institute of Technology}\\
%{\tt tu.zhaopeng@huawei.com}		       &        {Shenzhen Graduate School}\\
%{\tt lu.zhengdong@huawei.com}                 &       {\tt baotianchina@gmail.com}\\
%{\tt hangli.hl@huawei.com}
%\end{tabular}
%}

\maketitle
\begin{abstract}
We propose a novel method for translation selection in statistical machine translation, in which a convolutional neural network is employed to judge the similarity between a phrase pair in two languages. The specifically designed convolutional architecture encodes not only the semantic similarity of the translation pair, but also the context containing the phrase in the source language. Therefore, our approach is able to capture {\em context-dependent} semantic similarities of translation pairs. We adopt a curriculum learning strategy to train the model: we classify the training examples into easy, medium, and difficult categories, and gradually build the ability of representing phrase and sentence level context by using training examples from easy to difficult. Experimental results show that our approach significantly outperforms the baseline system by up to 1.4 BLEU points.
\end{abstract}

\section{Introduction}

In a conventional statistical machine translation (SMT)  system, the translation model is constructed in two steps~\cite{Koehn:2003}. 
First, bilingual phrase pairs respecting to the word alignments are extracted from a word-aligned parallel corpus. 
Second, the phrase pairs are assigned with scores calculated using their relative frequencies in the same corpus. 
However, only finding and utilizing translation pairs based on their surface forms is not sufficient: the conventional approach often fails to capture translation pairs which are grammatically and semantically similar.

To alleviate the above problems, several researchers have proposed learning and utilizing semantically similar translation pairs  in a continuous space~\cite{Gao:2014:ACL,Zhang:2014:ACL,Cho:2014:EMNLP}. The core idea is that the two phrases in a translation pair should share the same semantic meaning and have similar (close) feature vectors in the continuous space. A matching score is computed by measuring the distance between the feature vectors of the phrases, and is incorporated into the SMT system as an additional feature.

The above methods, however, {\em neglect the information of local contexts}, which has been proven to be useful for disambiguating translation candidates during decoding~\cite{He:2008:COLING,Marton:2008:ACL}.
The matching scores of translation pairs are treated the same, even they are in different contexts. 
Accordingly, the methods fail to adapt to local contexts and lead to precision issues for specific sentences in different contexts.

To capture useful context information, we propose a convolutional neural network architecture to measure context-dependent semantic similarities between phrase pairs in two languages.
For each phrase pair, we use the sentence containing the phrase in source language as the context.
With the convolutional neural network, we summarize the information of a phrase pair and its context, and further compute the pair's matching score with a multi-layer perceptron. 

We discriminately train the model using a curriculum learning strategy.
We classify the training examples (i.e. triples (source phrase with its context, positive candidate, negative candidate)) according to the difficulty level of distinguishing the positive candidate (i.e. correct translation for the source phrase in the specific context) from the negative candidate (i.e. a bad translation in this context).
Then we train the model to learn the semantic information from {\em easy} ({basic semantic similarities between phrase pairs}) to {\em difficult} ({context-dependent semantic similarities}).

Experimental results on a large-scale translation task show that the
context-dependent convolutional matching (CDCM) model improves the performance by up to 1.4 BLEU points over a strong phrase-based SMT system.
Moreover, the CDCM model significantly outperforms its context-independent counterpart, proving that it is necessary to incorporate local contexts into SMT.

\noindent{\bf \em Contributions.} Our key contributions include:
\begin{itemize}
 \item we introduce a novel CDCM model to capture context-dependent semantic similarities between phrase pairs (Section~\ref{section-CDCM-model});
 \item we develop a novel learning algorithm to train the CDCM model using a curriculum learning strategy (Section~\ref{section-training}).
\end{itemize}

\section{Related Work}
  
Our research builds on previous work in the field of context-dependent rule matching and bilingual phrase representations.

There is a line of work that employs local contexts over discrete representations of words or phrases. For example,~\namecite{He:2008:COLING}, ~\namecite{Liu:2008:EMNLP} and ~\namecite{Marton:2008:ACL} employed within-sentence contexts that consist of discrete words to guide rule matching. However, these discrete context features usually suffer the data sparseness problem.
In addition, these models treated each word as a distinct feature, which can not leverage the semantic similarity between words as our model.
~\namecite{Wu:2014:EMNLP} exploited discrete contextual features in the source sentence (e.g. words and part-of-speech tags) to learn better bilingual word embeddings for SMT. However, they only focused on frequent phrase pairs and induced phrasal similarities by simply summing up the matching scores of all the embraced words. In this study, we take into account all the phrase pairs and directly compute phrasal similarities with convolutional representations of the local contexts, integrating the strengths associated with the convolutional neural networks~\cite{Collobert:2008:ICML}.

Another line of work focuses on capturing the document-level contexts via distributed representations.
For instance,~\namecite{Xiao:2012:ACL} and~\namecite{Cui:2014:ACL} incorporated document-level topic information to select more semantically matched rules. 
Although many sentences share the same topic with the document where they occur, there are a lot of sentences actually do have topics different from those of their documents~\cite{Xiong:2013:AAAI}.
While these general contexts over the whole document may be not precise enough for the specific sentences in contexts different from the document,
 our approach is capable of learning the representations for different sentences respectively. 
Moreover, they learned distributed representations for documents rather than phrases and derived distributed phrase representations from the corresponding documents, while we attempt to build and train a single, large neural network that reads phrase pairs with contexts and outputs the match degrees directly.

In recent years, there has also been growing interest in bilingual phrase representations that group phrases with a similar meaning across different languages.
Based on that translation equivalents share the same semantic meaning, they can supervise each other to learn their semantic phrase embeddings in a continuous space.
For example,~\namecite{Gao:2014:ACL} projected phrases from both source and target sides into a common, continuous space that is language independent. Although~\namecite{Zhang:2014:ACL} did not enforce the phrase embeddings from both sides to be in the same continuous space, they exploited a transformation between the two semantic embedding spaces.
However, these models focused on capturing semantic similarities between phrase pairs in the global contexts, and neglected the local contexts, thus ignored the useful discriminative information.
Alternatively, we integrate the local contexts into our convolutional matching architecture to obtain context-dependent semantic similarities.

~\namecite{Meng:2015:ACL} and ~\namecite{Zhang:2015:IJCAI} have proposed independently to summary source sentences with convolutional neural networks. However, they both extend the neural network joint model (NNJM) of~\namecite{Devlin:2014:ACL} to include the whole source sentence, while we focus on capturing context-dependent semantic similarities of translation pairs.

\section{Context-Dependent Convolutional Matching Model}
\label{section-CDCM-model}

\begin{figure*}[!tb]
\centering
\includegraphics[width=\textwidth]{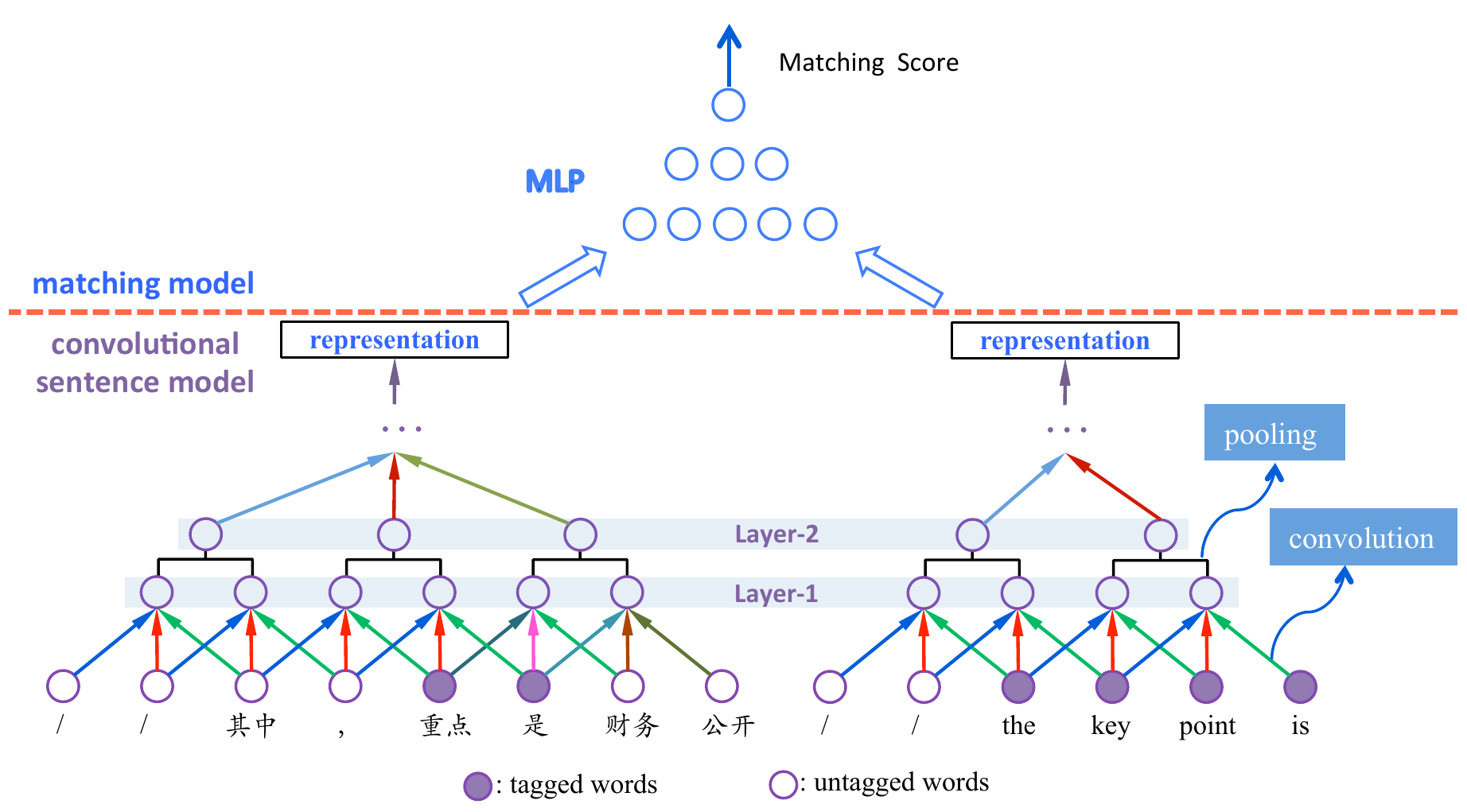}
\caption{Architecture of the CDCM model. The {\em convolutional sentence model} (bottom) summarizes the meaning of the tagged sentence and target phrase, and the {\em matching model} (top) compares the representations using a multi-layer perceptron. The symbol ``/'' indicates the all-zero padding turned off by the gating function.}
\label{figure-matching}
\end{figure*}

The model architecture, shown in Figure~\ref{figure-matching}, is a variant of the convolutional architecture of~\namecite{Hu:2014:NIPS}. It consists of two components:
\begin{itemize}
  \item {\em convolutional sentence model} that summarizes the meaning of the source sentence and the target phrase;
  \item {\em matching model} that compares the two representations with a multi-layer perceptron~\cite{Bengio:2009:FTML}.
\end{itemize}

Let $\hat{e}$ be a target phrase and ${\bf f}$ be the source sentence that contains the source phrase aligning to $\hat{e}$. We first project ${\bf f}$ and $\hat{e}$ into feature vectors ${\bf x}$ and ${\bf y}$ via the convolutional sentence model, and then compute the matching score $s({\bf x}, {\bf y})$ by the matching model. Finally, the score is introduced into a conventional SMT system as an additional feature.

\noindent{\bf Convolutional sentence model.} As shown in Figure~\ref{figure-matching}, the model takes as input the embeddings of words (trained beforehand elsewhere) in ${\bf f}$ and $\hat{e}$. It then iteratively summarizes the meaning of the input through layers of convolution and pooling, until reaching a fixed length vectorial representation in the final layer.

In Layer-1, the convolution layer takes sliding windows on ${\bf f}$ and $\hat{e}$ respectively, and models all the possible compositions of neighbouring words. The convolution involves a {\em filter} to produce a new feature for each possible composition. Given a $k$-sized sliding window $i$ on ${{\bf f}}$ or ${\hat{e}}$, for example, the $j$th convolution unit of the composition of the words is generated by:
\begin{equation}
{{\bf c}_i}^{(1,j)} = g(\hat{{\bf c}_i}^{(0)})\cdot\phi({\bf w}^{(1,j)}\cdot \hat{{\bf c}_i}^{(0)} + {\bf b}^{(1,j)})
\end{equation}
where
\begin{itemize}
  \item $g(\cdot)$ is the gate function that determines whether to activate $\phi(\cdot)$;
  \item $\phi(\cdot)$ is a non-linear activation function. In this work, we use ReLu~\cite{Dahl:2013:ICASSP} as the activation function;
  \item ${\bf w}^{(1,j)}$  is the parameters for the $j$th convolution unit on Layer-1, with matrix ${\bf W}^{(1)}=[{\bf w}^{(1,1)}, \dots, {\bf w}^{(1,J)}]$;
  \item $\hat{{\bf c}_i}^{(0)}$ is a vector constructed by concatenating word vectors in the $k$-sized sliding widow $i$;
  \item ${\bf b}^{(1,j)}$ is a bias term, with vector ${\bf B}^{(1)}=[{\bf b}^{(1,1)}, \dots, {\bf b}^{(1,J)}]$.
\end{itemize}

To distinguish the phrase pair from its context, we use one additional dimension in word embeddings: $1$ for words in the phrase pair and $0$ for the others.
After transforming words to their tagged embeddings, the convolutional sentence model takes multiple choices of composition using sliding windows in the convolution layer. 
Note that sliding windows are allowed to cross the boundary of the source phrase to exploit both phrasal and contextual information.

In order to avoid the length variability of source sentences and target phrases, we add all-zero paddings at the end of the source sentence and target phrase until their maximum length. Moreover, we use the gate function $g(\cdot)$ to eliminate the effect of the all-zero padding by setting output vector to all-zeros if the input is all-zeros.

In Layer-2, we apply a local max-pooling in non-overlapping $1\times2$ windows for every convolution unit
\begin{equation}
{\bf c}^{(2,j)}_{i} = \max\{{\bf c}^{(1,j)}_{2i},{\bf c}^{(1,j)}_{2i+1}\}
\end{equation}
In Layer-3, we perform convolution on output from Layer-2:
\begin{equation}
{{\bf c}_i}^{(3,j)} = g(\hat{{\bf c}_i}^{(2)})\cdot\phi({\bf w}^{(3,j)}\cdot \hat{{\bf c}_i}^{(2)} + {\bf b}^{(3,j)})
\end{equation}
After more convolution and max-pooling operations, we obtain two feature vectors for the source sentence and the target phrase, respectively.

\noindent{\bf Matching model.}
The matching score of a source sentence and a target phrase can be measured as the similarity between their feature vectors.
Specifically, we use the multi-layer perceptron (MLP), a nonlinear function for similarity, to compute their matching score. First we use one layer to combine their feature vectors to get a hidden state $h_c$.
\begin{equation}
h_c = \phi(w_{c}\cdot[{\bf x}_{\bar{f}_i}:{\bf y}_{\bar{e}_j}] + b_{c})
\label{equation-hc}
\end{equation}
Then we get the matching score from the MLP:
\begin{equation}
s({\bf x}, {\bf y}) = MLP(h_c)
\label{equation-match-score}
\end{equation}

\section{Training}
\label{section-training}

Ideally, the trained CDCM model is expected to assign a higher matching score to a positive example (a source phrase in a specific context ${\bf f}$ and its correct translation ${\hat{e}}^+$), and a lower score to a negative example (the source phrase and a bad translation ${\hat{e}}^-$ in the specific context). To this end, we employ a discriminative training strategy with a max-margin objective.

Suppose we are given the following triples (${\bf x}, {\bf y}^+, {\bf y}^-$) from the oracle, where ${\bf x}, {\bf y}^+, {\bf y}^-$  are the feature vectors for ${\bf f}, {\hat{e}}^+, {\hat{e}}^-$ respectively.
We have the ranking-based loss as objective:
\begin{equation}
L_\Theta({\bf x}, {\bf y}^+, {\bf y}^-) = \max(0, 1+s({\bf x}, {\bf y}^-)-s({\bf x}, {\bf y}^+))
\label{equation-objective}
\end{equation}
where $s({\bf x}, {\bf y})$ is the matching score function defined in Eq.~\ref{equation-match-score}, $\Theta$ consists of parameters for both the convolutional sentence model and MLP.
The model is trained by minimizing the above objective, to encourage the model to assign higher matching scores to positive examples and to assign lower scores to negative examples.
We use stochastic gradient descent (SGD) to optimize the model parameters $\Theta$.

Note that the CDCM model aims at capturing contextual representations that can distinguish good translation candidates from bad ones in various contexts. To this end, we propose a two-step approach. First, we initialize the model with context-dependent bilingual word embeddings to start with strong contextual and semantic equivalence at the word level (Section~\ref{section-bilingual-word-embedding}).
Second, we train the CDCM model with a curriculum strategy to learn the context-dependent semantic similarity at the phrase level from {\em easy} ({basic semantic similarities between the source and target phrase pair}) to {\em difficult} ({context-dependent semantic similarities for the same source phrase in varying contexts}) (Section~\ref{section-curriculum-training}).

\subsection{Initialization by Context-Dependent Bilingual Word Embeddings}
\label{section-bilingual-word-embedding}

\begin{figure}[!tb]
\centering
\includegraphics[width=0.7\textwidth]{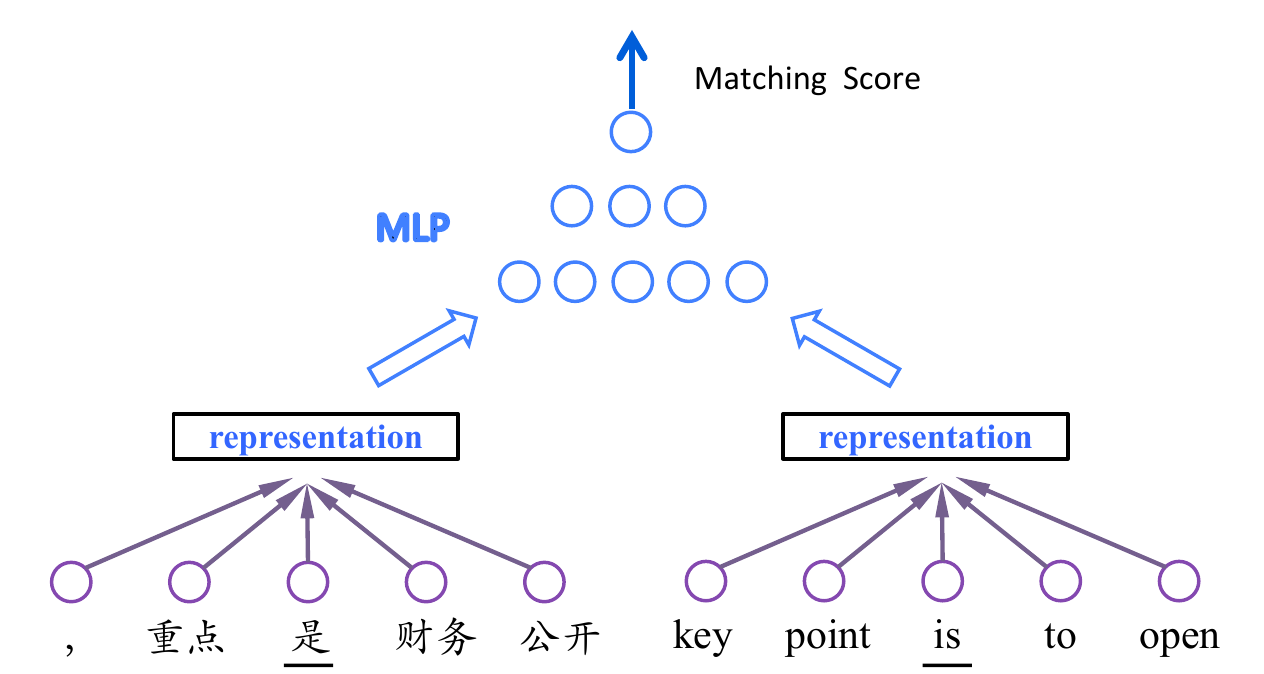}
\caption{Architecture of the CDCM bilingual word embedding model. }
\label{figure-bilingual-embeddings}
\end{figure}

Model initialization plays a critical role in a non-convex problem.
The initialization of the CDCM model is the embeddings of words on both languages, a real-value and dense representation of words.
Typical word embeddings are trained on monolingual data~\cite{Mikolov:2013:ArXiv}, thus fails to capture the useful semantic relationship across languages.
It has been shown that bilingual word embeddings represent a substantial step in better capturing semantic equivalence at the word level~\cite{Zou:2013:EMNLP,Wu:2014:EMNLP}, thus could initialize our model with strong semantic information.
Bilingual word embeddings refer to the semantic embeddings associated across two languages so that similar units in each language and across languages have similar representations.
~\namecite{Zou:2013:EMNLP} utilized MT word alignments to encourage pairs of frequently aligned words to have similar word embeddings, while~\namecite{Wu:2014:EMNLP} improved bilingual word embeddings with discrete contextual information.

Inspired by the above studies, we propose a context-dependent bilingual word embedding model that exploits both the word alignments and contextual information, as shown in Figure~\ref{figure-bilingual-embeddings}.
Given an aligned word pair ($f_i$, $e_j$), the context is extracted from the nearby window on each side (the left two words and the right two words in this work).
Let $\bar{f}_i=f_{i-2},f_{i-1},f_i,f_{i+1},f_{i+2}$ and $\bar{e}_j=e_{j-2},e_{j-1},e_j,e_{j+1},e_{j+2}$ be the contextual sequence for the above word pair. We get their vectorial representations by:
\begin{eqnarray}
{\bf x}_{\bar{f}_i} = \phi({\bf w}_f\cdot Le(\bar{f_i})+{\bf b}_f) \\
{\bf y}_{\bar{e}_j} = \phi({\bf w}_e\cdot Le(\bar{e_j})+{\bf b}_e)
\end{eqnarray}
where $Le(\cdot)$ converts word sequences into embeddings and returns a vector by concatenating the embeddings.

Similarly, we calculate matching score for ${\bf x}_{\bar{f}_i}$ and ${\bf y}_{\bar{e}_j}$ according to Eq.~\ref{equation-hc} and Eq.~\ref{equation-match-score}. The bilingual word embedding model is also trained by minimizing the objective in Eq.~\ref{equation-objective}.
The negative examples are constructed by replacing either $f_i$ or $e_j$ with words randomly chosen from the corresponding vocabulary.

\subsection{Curriculum Training}
\label{section-curriculum-training}

Curriculum learning, first proposed by~\namecite{Bengio:2009:ICML} in machine learning, refers to a sequence of training strategies that start small, learn easier aspects of the task, and then gradually increase the difficulty level. It has been shown that the curriculum learning can benefit the non-convex training by giving rise to improved generalization and faster convergence.
The key point is that the training examples are not randomly presented but organized in a meaningful order which illustrates gradually more concepts, and gradually more complex ones.

For each positive example (${\bf f}, \hat{e}^+$), we have three types of negative examples according to the difficulty level of distinguishing the positive example from them:
\begin{itemize}
  \item {\em Easy}: target phrases randomly chosen from the phrase table;
  \item {\em Medium}: target phrases extracted from the aligned target sentence for other non-overlap source phrases in the source sentence;
  \item {\em Difficult}: target phrases extracted from other candidates for the same source phrase.
\end{itemize}
We want the CDCM model to learn the following semantic information from easy to difficult:
\begin{itemize}
  \item  the {\em basic semantic similarity} between the source sentence and target phrase from the {\em easy} negative examples;
  \item  the {\em general semantic equivalent} between the source and target phrase pair from the {\em medium} negative examples;
  \item  the {\em context-dependent semantic similarities} for the same source phrase in varying contexts from the {\em difficult} negative examples.
\end{itemize}

\begin{algorithm}[t]
\small
\begin{algorithmic}[1]
\Procedure {CURRICULUM-TRAINING}{$\mathcal{T}$, $W$}
\State {{$N_1$} $\leftarrow$ {\em easy\_negative}($\mathcal{T}$)}
\State {{$N_2$} $\leftarrow$ {\em medium\_negative}($\mathcal{T}$)}
\State {{$N_3$} $\leftarrow$ {\em difficult\_negative}($\mathcal{T}$)}
\State {$T$ $\leftarrow$ $N_1$}
\State {{CURRICULUM($T$, $n\cdot t$)}	\Comment{\em CUR. easy}}
\State {{$T$} $\leftarrow$ MIX([$N_1$, $N_2$])}
\State {{CURRICULUM($T$, $n\cdot t$)}	\Comment{\em CUR. medium}}
 \For {$step$ $\leftarrow$ $1\dots n$}
\State {{$T$} $\leftarrow$ MIX([$N_1$, $N_2$, $N3$], step)}
\State {{CURRICULUM($T$, $t$)}	\Comment{\em CUR. difficult}}
\EndFor
\EndProcedure
\Procedure {CURRICULUM}{{$T$, $K$}}
\State {\em iterate until reaching a local minima or K iterations}
\State {\em calculate $L_\Theta$ for a random  instance in $T$}
\State {$\Theta = \Theta - \eta \cdot \frac{\partial L_\Theta}{\partial \Theta}$ \Comment{\em update parameters}}
\State {$W = W - \eta \cdot 0.01 \cdot \frac{\partial L_\Theta}{\partial W}$ \Comment{\em update embeddings}}
\EndProcedure
\Procedure {MIX}{$\bf N$, $s=0$}
 \State {$len$ $\leftarrow$ length of $\bf N$}
 \If {$len < 3$}
 \State{$T$ $\leftarrow$ sampling with $[0.5, 0.5]$ from N}
 \Else
 \State{$T$ $\leftarrow$ sampling with $[\frac{1}{s+2}, \frac{1}{s+2}, \frac{s}{s+2}]$ from N}
 \EndIf
\EndProcedure
\end{algorithmic}
\caption{Curriculum training algorithm. Here $\mathcal{T}$ denotes the training examples, $W$ the initial word embeddings, $\eta$ the learning rate in SGD, $n$ the pre-defined number, and $t$ the number of training examples.}
\label{algorithm-curriculum-learning}
\end{algorithm}

Alg.~\ref{algorithm-curriculum-learning} shows the curriculum training algorithm for the CDCM model. We use different portions of the overall training instances for different curriculums (lines 2-11). For example,  we only use the training instances that consist of positive examples and {\em easy} negative examples in the {\em easy} curriculum (lines 5-6). For the latter curriculums, we gradually increase the difficulty level of the training instances (lines 7-12).

For each curriculum (lines 12-16), we compute the gradient of the loss objective $L_\Theta$ and learn $\Theta$ using the SGD algorithm. Note that we meanwhile update the word embeddings to better capture the semantic equivalence across languages during training. If the loss function $L_\Theta$ reaches a local minima or the iterations reach the pre-defined number, we terminate this curriculum.

\section{Experiments}
\label{section-experiments}

In this section, we try to answer two questions:
\begin{itemize}
  \item[1] Does the proposed approach achieve higher translation quality than the baseline system? 
  Does the approach outperform its context-independent counterpart?
  \item[2] Does model initialization by  bilingual word embeddings outperforms its monolingual counterpart in terms of translation quality?
\end{itemize}

In Section~\ref{section-evaluation-of-translation-quality}, we evaluate our approach on a Chinese-English  translation task. 
By using the CDCM model, our approach achieves significant improvement in BLEU score by up to 1.4 points.
Moreover, the CDCM model significantly outperforms its context-independent counterpart, confirming our hypothesis that local contexts are very useful for machine translation.

In Section~\ref{section-evaluation-of-word-embedding}, we compare model initializations by bilingual word embeddings and by conventional monolingual word embeddings. Experimental results show that the initialization by bilingual word embeddings outperforms its monolingual counterpart consistently, indicating that bilingual word embeddings give a better initialization of the CDCM model.

\subsection{Setup}

We carry out our experiments 
on the NIST Chinese-English translation tasks. Our training data contains 1.5M sentence pairs coming from LDC dataset. The corpus includes LDC2002E18, LDC2003E07, LDC2003E14, Hansards portion of LDC2004T07, LDC2004T08 and LDC2005T06.
We train a 4-gram language model on the Xinhua portion of the GIGAWORD corpus using the SRI Language Toolkit~\cite{Stolcke:2002}.
We use the 2002 NIST MT evaluation test data as the development data, and the 2004, 2005 NIST MT evaluation test data as the test data.
We use minimum error rate training~\cite{Och:2003b} to optimize the feature weights. For evaluation, case-insensitive NIST BLEU~\cite{Papineni:2002} is used to measure translation performance.

For training the neural networks, we use 4 convolution layers for source sentences and 3 convolution layers for target phrases. For both of them, 4 pooling layers (pooling size is 2) are used, and all the feature maps are 100.
We set the sliding window $k=3$, and the learning rate $\eta=0.02$. All the parameters are selected based on the development data.
To produce high-quality bilingual phrase pairs to train the CDCM model, we perform forced decoding on the bilingual training sentences and collect the used phrase pairs.
We obtain 2.4M unique phrase pairs (length ranging from 1 to 7) and 20.2M phrase pairs in different contexts. Since the curriculum training in the CDCM model requires that each source phrase should have at least two corresponding target phrases, we obtain 13.5M phrase pairs after we remove the undesirable ones.

\subsection{Evaluation of Translation Quality}
\label{section-evaluation-of-translation-quality}

\begin{table}
\centering
\begin{tabular}{llll}
    \hline
    Models       &  MT04  &  MT05  &  All\\
    \hline
    Baseline       &  34.86  &  33.18  &  34.40\\
    \hline
    CICM           &   35.82$^\alpha$  & 33.51$^\alpha$   &  34.95$^\alpha$\\
   \hline
    CDCM$_1$        &  35.87$^\alpha$    &  33.58	       &   35.01$^\alpha$\\
    CDCM$_2$        &  35.97$^\alpha$    &  33.80$^\alpha$   &   35.21$^\alpha$\\
    CDCM$_3$        &  36.26$^{\alpha\beta}$    &   33.94$^{\alpha\beta}$  &   35.40$^{\alpha\beta}$\\
    \hline
\end{tabular}
\caption{Evaluation of translation quality. CDCM$_k$ denotes the CDCM model trained in the $k$th curriculum, CICM denotes its context-independent counterpart, and ``All'' is the combined test sets. The superscripts $\alpha$ and $\beta$ indicate statistically significant difference ($p<0.05$) from Baseline and CICM, respectively.}
\label{table-translation-results}
\end{table}

\begin{figure*}[!tb]
\centering
\includegraphics[width=\textwidth]{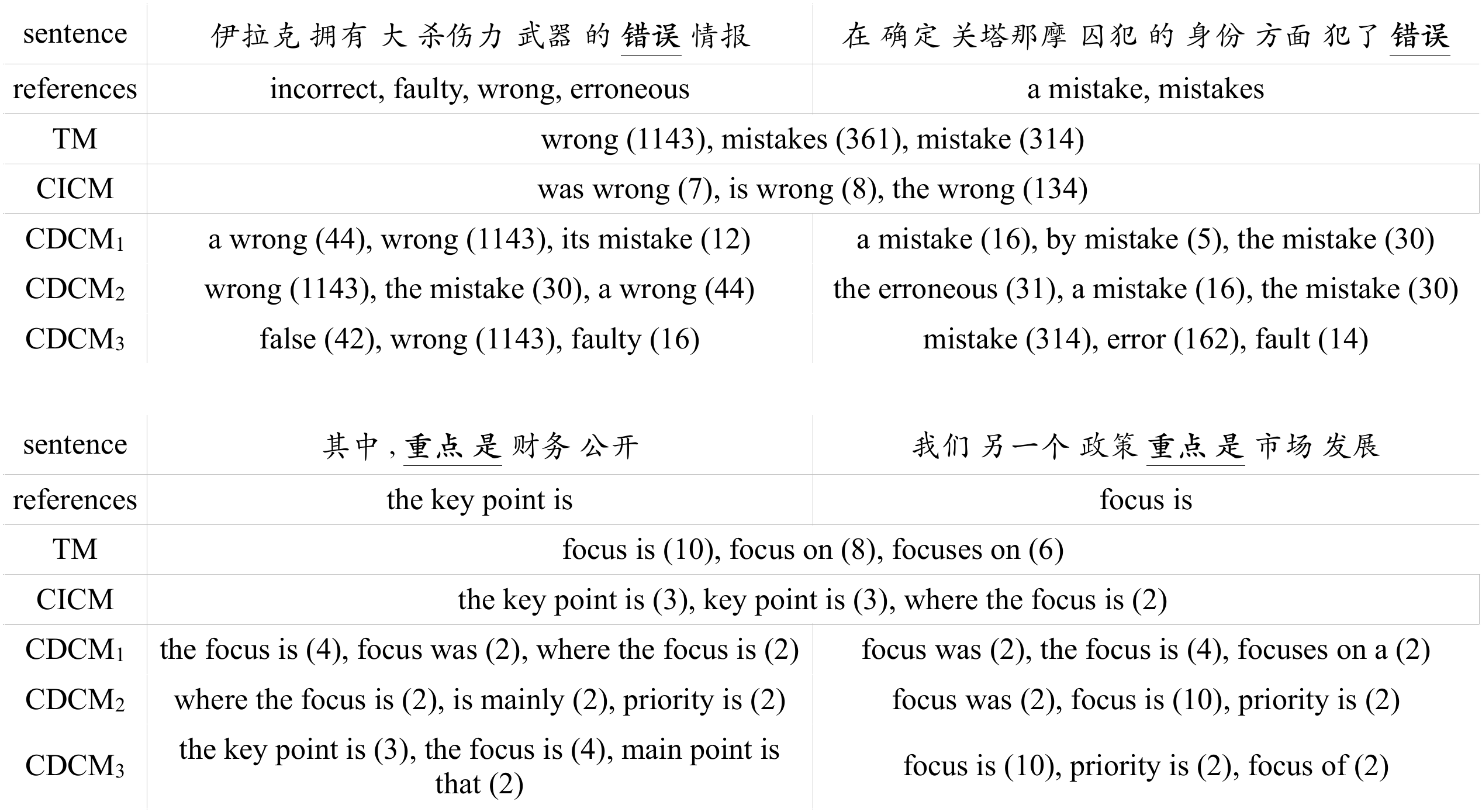}
\caption{The top ranked target phrases according to the translation model (TM) and the CDCM model. The number in the bracket is the co-occurrence of the source-target phrase pair in the corpus.}
\label{figure-cases}
\end{figure*}

We have two baseline systems:
\begin{itemize}
  \item {\em Baseline}: The baseline system is an open-source system of the phrase-based model -- Moses~\cite{Koehn:ACL:2007} with a set of common features, including translation models, word and phrase penalties, a linear distortion model, a lexicalized reordering model, and a language model. 
  \item CICM (context-{\em independent} convolutional matching) model: Following the previous works~\cite{Gao:2014:ACL,Zhang:2014:ACL,Cho:2014:EMNLP}, we calculate the matching degree of a phrase pair without considering any contextual information. Each unique phrase pair serves as a positive example and a randomly selected target phrase from the phrase table is the corresponding negative example. 
  The matching score is also introduced into Baseline as an additional feature.
\end{itemize}

Table~\ref{table-translation-results} summaries the results of CDCMs trained from different curriculums. No matter from which curriculum it is trained, the CDCM model significantly improves the translation quality on the overall test data (with gains of 1.0 BLEU points).
The best improvement can be up to 1.4 BLEU points on MT04 with the fully trained CDCM.
As expected, the translation performance is consistently increased with curriculum growing. This indicates that the CDCM model indeed captures the desirable semantic information by the curriculum learning from easy to difficult.

Comparing with its context-independent counterpart (CICM, Row 2), the CDCM model shows significant improvement on all the test data consistently. We contribute this to the incorporation of useful discriminative information embedded in the local context. In addition, the performance of CICM is comparable with that of CDCM$_1$. This is intuitive, because both of them try to capture the basic semantic similarity between the source and target phrase pair.

\noindent{\bf \em Qualitative Analysis.} Figure~\ref{figure-cases} lists some interesting cases to show why the CDCM model improves the performance.
We analyze the phrase pair scores computed by the CDCM model against the phrase translation probabilities from the translation model. 
First, the CDCM model scores phrase pairs based rather on the semantic similarity and the contextual information than on their co-occurrences in the corpus. Therefore, it is complementary to the translation model.
Second, with the growing of curriculum, our model is more likely to capture the context-dependent semantic similarities between phrase pairs.
In most cases, the choices of translation candidates by the fully trained CDCM model (i.e. CDCM$_3$) are closer to actual translations for both frequent and less frequent phrases.
Third, though the CICM model captures the semantic similarities between phrase pairs, it fails to adapt to different local contexts as well. In contrast, the CDCM model is able to provide different translation candidates based on the discriminative information embedded in the local contexts.

\subsection{Evaluation of Bilingual Word Embeddings}
\label{section-evaluation-of-word-embedding}

\begin{table*}
\centering
\begin{tabular}{c|ccc|ccc}
    \hline
    \multirow{2}{*}{Models}       &  \multicolumn{3}{c|}{Monolingual} &
  \multicolumn{3}{c}{Bilingual}\\
                                                   &   MT04  &  MT05  &  All &  MT04  &  MT05  &  All\\
    \hline
    CDCM$_1$        &    35.74    &	33.38   &   34.85   &  35.87      &  33.58     &   35.01\\
    CDCM$_2$        &    35.80    &  33.59   &   35.04   &  35.97      &  33.80     &   35.21\\
    CDCM$_3$        &    35.95    &  33.65   &   35.14   &  36.26      &  33.94     &   35.40\\
    \hline
\end{tabular}
\caption{Comparison of the monolingual word embeddings ({\em Monolingual}) and the bilingual word embeddings ({\em Bilingual}) in terms of translation quality.}
\label{table-word-embedding}
\end{table*}

\begin{figure*}[!tb]
\centering
\includegraphics[width=\textwidth]{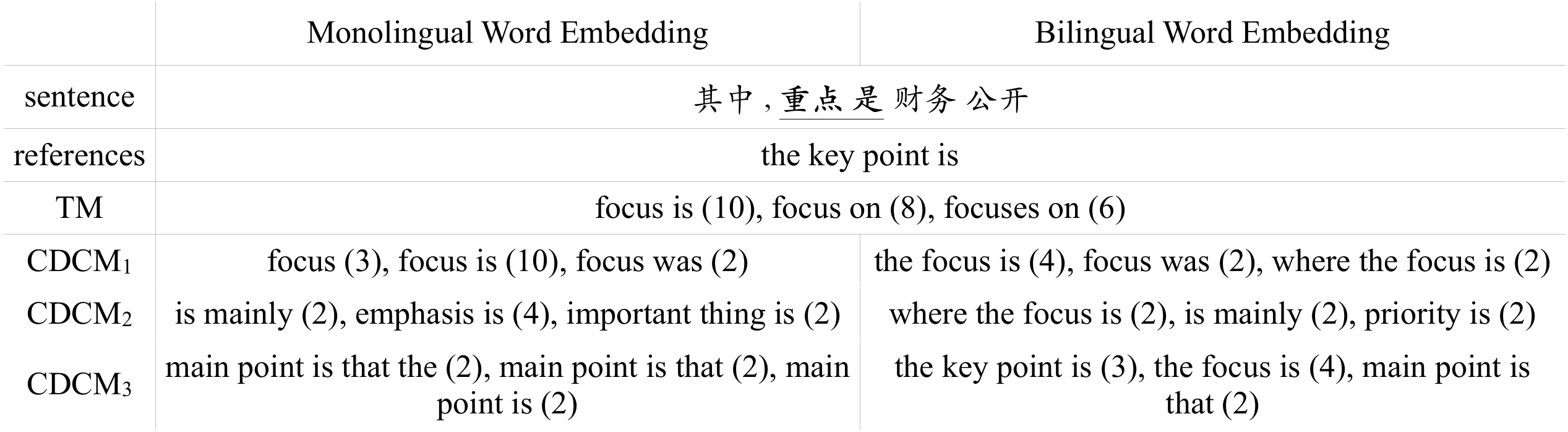}
\caption{The top ranked target phrases according to the CDCM models with different initilizations.}
\label{figure-bilingual-cases}
\end{figure*}

In this section, we will investigate the influence of the bilingual word embeddings we use to initialize the CDCM model.
We use the Word2Vec~\cite{Mikolov:2013:ArXiv} to train  the monolingual word embeddings. 
We train the bilingual word embeddings using the approach described in Section~\ref{section-bilingual-word-embedding}. Dimensions of both bilingual and monolingual embeddings are 50.

Table~\ref{table-word-embedding} shows the comparative results between bilingual and monolingual word embeddings. As seen, our bilingual word embedding model outperforms its monolingual counterpart consistently.~\namecite{Zou:2013:EMNLP} and~\namecite{Wu:2014:EMNLP} reported that word-level semantic relationships across languages, captured by the bilingual word embeddings, boost machine translation performance.
Our results reconfirm these findings.

\noindent{\bf \em Qualitative Analysis.} Figure~\ref{figure-bilingual-cases} lists some cases to show why the context-dependent bilingual word embeddings produce consistent improvements. As seen, the CDCM model initialized by bilingual word embeddings produces more discriminative results than its monolingual counterpart. Take CDCM$_3$ as an example, the monolingual word embeddings scenario prefers the candidates that contain ``{\em main point is}", while its bilingual counterpart selects different candidates that share the same semantic meaning. One possible reason is that bilingual and contextual information helps to capture the semantic relationships between words across languages~\cite{Yang:2013:ACL}, thus better phrasal similarities by using principle of compositionality.

\subsection{Discussion}

\noindent{\bf \em Convolutional Model vs. Recursive Model.}
Previous works on bilingual phrase representations usually employ Recurrent Neural Network (RNN)~\cite{Cho:2014:EMNLP} or Recursive AutoEncoder (RAE)~\cite{Zhang:2014:ACL}. It has been observed in~\cite{Kalchbrenner:2013:EMNLP,Sutskever:2014:NIPS,Cho:2014:SSST} that the recursive approaches suffer from a significant drop in translation quality when translating long sentences. In contrast, ~\namecite{Kalchbrenner:2014:ACL} show that the convolutional model could represent the semantic content of a long sentence accurately.
Therefore, we choose the convolutional architecture to model the meaning of sentence.

\noindent{\bf \em Limitations.}
Unlike recursive models, the convolutional architecture has a fixed depth, which bounds the level of composition. In this task, this limitation can be largely compensated with a network afterwards that can take a ``global'' synthesis on the learned sentence representation.

One of the hypotheses we tested in the course of this research was disproved. We thought it likely that the {\em difficult} curriculum (i.e. distinguish the correct translation from other candidates for a given context) would contribute most to the improvement, since this circumstance is more consistent with the real decoding procedure. This turned out to be false, as shown in Table~\ref{table-translation-results}. One possible reason is that the ``negative'' examples (other candidates for the same source phrase) may share the same semantic meaning with the positive one, thus give a wrong guide in the supervised training. Constructing a reasonable set of negative examples that are more semantically different from the positive one is left for our future work.

\section{Conclusion}

In this paper, we propose a context-dependent convolutional matching model to capture semantic similarities between phrase pairs that are sensitive to contexts. Experimental results show that our approach significantly improves the translation performance and obtains improvement of 1.0 BLEU scores on the overall test data.

Integrating deep architecture into context-dependent translation selection is a promising way to improve machine translation.
This paper is the first step in what we hope will be a long and fruitful journey.
In the future, we will try to exploit contextual information at the target side  ({\em e.g.,} partial translations).

\bibliographystyle{acl}
%\balance
\bibliography{all}

\end{document}